\documentclass[final,5p,times,twocolumn]{elsarticle}

\usepackage{amssymb}
\usepackage{amsmath}

\usepackage{csquotes}
\usepackage{amsmath}
\usepackage{amssymb}
\usepackage{multirow}
\usepackage{tabularx}
\usepackage{tcolorbox}
\usepackage{booktabs}

\journal{Neurocomputing}

\begin{document}

\begin{frontmatter}



\title{Leveraging Inter-Chunk Interactions for Enhanced Retrieval in Large Language Model-Based Question Answering}


\author[NEU,NSAI]{Tiezheng Guo}
\ead{2310725@stu.neu.edu.cn}

\author[NEU,NSAI]{Chen Wang}
\ead{wangchen-neu@neusoft.com}

\author[NEU]{Yanyi Liu}
\ead{2290175@stu.neu.edu.cn}

\author[NEU]{Jiawei Tang}
\ead{2301944@stu.neu.edu.cn}

\author[NEU,NSAI]{Pan Li}
\ead{li_pan@neusoft.com}

\author[NEU,NSAI]{Sai Xu}
\ead{xu_s@neusoft.com}

\author[NEU,NSAI]{Qingwen Yang}
\ead{yang_qw@neusoft.com}

\author[NSAI]{Xianlin Gao}
\ead{gaoxianlin@neusoft.com}

\author[NSAI]{Zhi Li}
\ead{zhi-li@neusoft.com}

\author[NEU,NSAI]{Yingyou Wen\corref{corresponding}}
\ead{wenyingyou@mail.neu.edu.cn}

\cortext[corresponding]{Corresponding author.}

\affiliation[NEU]{
            organization={School of Computer Science and Engineering},
            addressline={Northeastern University}, 
            city={Shenyang},
            postcode={110169},
            country={China}}
\affiliation[NSAI]{
            organization={Neusoft AI Magic Technology Research},
            city={Shenyang},
            postcode={110179},
            country={China}}

\begin{abstract}
  Retrieving external knowledge and prompting large language models with relevant information is an effective paradigm to enhance the performance of question-answering tasks. Previous research typically handles paragraphs from external documents in isolation, resulting in a lack of context and ambiguous references, particularly in multi-document and complex tasks. To overcome these challenges, we propose a new retrieval framework IIER, that leverages Inter-chunk Interactions to Enhance Retrieval. This framework captures the internal connections between document chunks by considering three types of interactions: structural, keyword, and semantic. We then construct a unified Chunk-Interaction Graph to represent all external documents comprehensively. Additionally, we design a graph-based evidence chain retriever that utilizes previous paths and chunk interactions to guide the retrieval process. It identifies multiple seed nodes based on the target question and iteratively searches for relevant chunks to gather supporting evidence. This retrieval process refines the context and reasoning chain, aiding the large language model in reasoning and answer generation. Extensive experiments demonstrate that IIER outperforms strong baselines across four datasets, highlighting its effectiveness in improving retrieval and reasoning capabilities.
\end{abstract}




\end{frontmatter}



\section{Introduction}

Large langugae models (LLM) have acquired superior reading comprehension and reasoning capabilities by pretraining on extensive natural langugae data \cite{Guo2023EvaluatingLL,li2023large}. They have demonstrated remarkable performance on a variety of tasks and benchmarks, particularly in the realm of question answering (QA) \cite{LIU2024127505, LUO2024128089}. Researchers are expanding the parameter scale of these models to enable them to retain more knowledge \cite{wu2023survey}. However, due to the absence of efficient methods to evaluate or edit their internalized knowledge \cite{hase2024does}, knowledge-intensive tasks remain a major challenge for LLMs \cite{xu2023search}. This issue is especially pronounced in specialized domain queries, where LLMs often struggle to generate factual information due to hallucinations \cite{manakul2023selfcheckgpt}.

Retrieval-Augmented Generation (RAG) is a paradigm that retrieves external knowledge sources to provide LLMs with relevant information before generating responses \cite{li2022survey}. This method effectively mitigates the impact of LLM's limited memory capacity regarding open-domain knowledge and significantly enhances the accuracy and factualness of their outputs. The core of this paradigm lies in retrieving supporting evidence from documents that contain the answers to the question. However, when dealing with complex multi-document question answering (MDQA) tasks, accurately understanding the question's constraints and covering all supporting evidence remains an open challenge \cite{Caciularu2023PeekAI,singh2021end}. This difficulty arises because previous research has treated the relationship between each text chunk and the target question in isolation. The retrieval models have concentrated solely on whether the main topic of each chunk aligns with the question \cite{ma2023query}. Imperfect preprocessing can lead to the incorrect truncation of continuous chunks. In the absence of context, anaphora and ambiguity within the chunk can have a significant impact on its original semantic expression \cite{uryupina2020annotating}. Consequently, retrieval models are limited by their accuracy of knowledge representation and similarity calculation, often only locating a vague range within the semantic space and failing to capture all valid supporting evidence. This phenomenon becomes more pronounced as the number of documents and the complexity of the question increase.

In multi-document tasks, the chunks relevant to the target question are typically correlated with each other. When they originate from the same source, these chunks are often adjacent in the original structure, providing contextual support for each other. When they come from different sources, they are usually connected together by common keywords or similar semantics. These interactions suggest potential parallel or progressive relationships in reasoning. By mining and representing these interactions, we can overcome the limitations of low similarity within the search scope. Retrieval models can leverage these connections to access additional supporting evidence, starting from highly relevant chunks. This approach enables retrieval to move beyond question-centered local searches, thereby improving the accuracy of evidence recall.

Based on these analyses, we propose a framework called IIER that uses \textbf{I}nter-chunk \textbf{I}nteractions to \textbf{E}nhance \textbf{R}etrieval. In this framework, all documents are divided into chunks, each treated as a node with text attributes. We identify three types of interactions between node pairs, including structural, semantic, and keyword interactions. These interactions are used as edge attributes linking the nodes. This approach enables IIER to transform all external documents into a unified Chunk-Interaction Graph (CIG) structure, effectively disregarding document differences. To leverage these interactions for retrieval, we design and fine-tune a graph-based evidence chain retriever. The retriever initiates the search from chunks that are easily accessible and highly relevant to the question. In each iteration, it utilizes the topology information to determine the optimal search direction, gradually moving closer to the supporting evidence needed for the question. Through iterative retrieval, IIER constructs multiple complete reasoning chains to assist LLM in question answering. Our contributions are as follows:

\begin{itemize}
\item We analyze the limitations of treating each chunk as an independent semantic fragment in RAG and propose a method to bridge related chunks by unifying multiple documents with a Chunk-Interaction Graph (CIG). In this approach, chunks are represented as nodes with text attributes, while the interactions—structural, semantic, and keyword—between these chunks are captured as edge attributes. This allows CIG to establish additional connections between chunks from various sources, thereby providing richer contextual information and overcoming the constraints of local context. 

\item  We design a graph-based evidence chain retriever that considers text attributes, previous retrieval information, and graph topology to guide the search direction within the CIG. We fine-tune the retriever to achieve adaptively search for each target question, enabling it to construct reasoning paths and context. This approach enhances the LLM's ability to reason about the target question.

\item We simulate real-world multi-document question answering scenarios by extending the QA datasets. In extensive experiments across all datasets, IIER achieves the best accuracy, demonstrating its effectiveness in enhancing retrieval and reasoning capabilities. Additionally, we conduct analysis experiments to validate the improvements introduced by CIG and evidence chain.
\end{itemize}

\section{Related Work}

\subsection{Retrieval-augmented LLM}

The emergence of large language models has brought revolutionary impact on many natural language processing tasks. However, LLMs still face significant limitations in various scenarios \cite{Gao2023RetrievalAugmentedGF}. Especially in QA domains that require specialized and real-time knowledge, the content generated by LLMs often suffers from issues such as factual inaccuracies or hallucinations \cite{Ji2022SurveyOH}. Retrieval-Augmented Generation has been proven to be an effective solution to these challenges. By retrieving evidence supporting the answer from external knowledge, RAG can enhance LLMs' reasoning and question answering performance \cite{Sun2023ThinkonGraphDA,Nie2022CapturingGS}. This paradigm typically consists of a text retriever and an evidence reader. The primary challenge in this domain is accurately identifying all the necessary textual information required to answer the question across various documents. To address the complexity challenge posed by long documents, RAPTOR \cite{Sarthi2024RAPTORRA} constructs texts into a tree structure at multiple granularities through text summarization, enhancing the accuracy of semantic similarity calculation. SANTA \cite{Li2023StructureAwareLM} enhances the sensitivity of the retriever in structured data by using structured-aware pre-training and masked entity prediction,  achieving better performance in understanding and searching structured data. In order to align retrieval and reasoning, REPLUG \cite{Shi2023REPLUGRB} treats the LLMs as black boxes and uses them as supervisory signals to guide the optimization of retrieval models, thereby making it possible to identify documents that help LLMs make better predictions. TOG\cite{Sun2023ThinkonGraphDA} and ROG\cite{Luo2023ReasoningOG}  deploy knowledge graphs as reliable knowledge sources and generate enhanced prompts by searching target entities in the graph.

\subsection{Multi-hop QA}

In challenging scenarios that require searching for pieces of supporting evidence from numerous documents and performing complex reasoning, multi-hop reasoning is widely deployed to enhance the accuracy and comprehensiveness of retrieval. In this task, a retrieval agent is typically used to iteratively select the optimal next hop direction and retrieve relevant information step by step. A common approach is to deploy LLM as the agent to guide the retrieval. HyKGE\cite{Jiang2023HyKGEAH}, KGP\cite{Wang2023KnowledgeGP} and ITER-RETGEN\cite{Shao2023EnhancingRL} utilize LLM to generate hypothesis outputs based on previous context, simulating the knowledge content that might be needed to complete the task. They compare the similarity between all potential evidence and the hypothesis output to identify the effective anchor points. However, these methods require instruction fine-tuning on LLMs to enhance their ability to analyze missing evidence, which incurs a high additional cost. In contrast, DecomP \cite{Khot2022DecomposedPA} and IRCoT \cite{Trivedi2022InterleavingRW} prompt LLM with CoT to decompose the task into subtasks, thereby improving the retrieval accuracy by optimizing these specific subtasks. They achieve the integration of LLM and external knowledge sources as plug-and-play modules, reducing the overhead of transfer. However, these methods often overlook the inherent connections between relevant chunks, making it difficult to handle tasks with a large number of chunks or ambiguous semantic expressions. Consequently, they still face challenges in effectively retrieving all supporting evidence and maintaining coherent reasoning logic.

\begin{figure*}[h]
    \centering
    \includegraphics[width=0.8\textwidth]{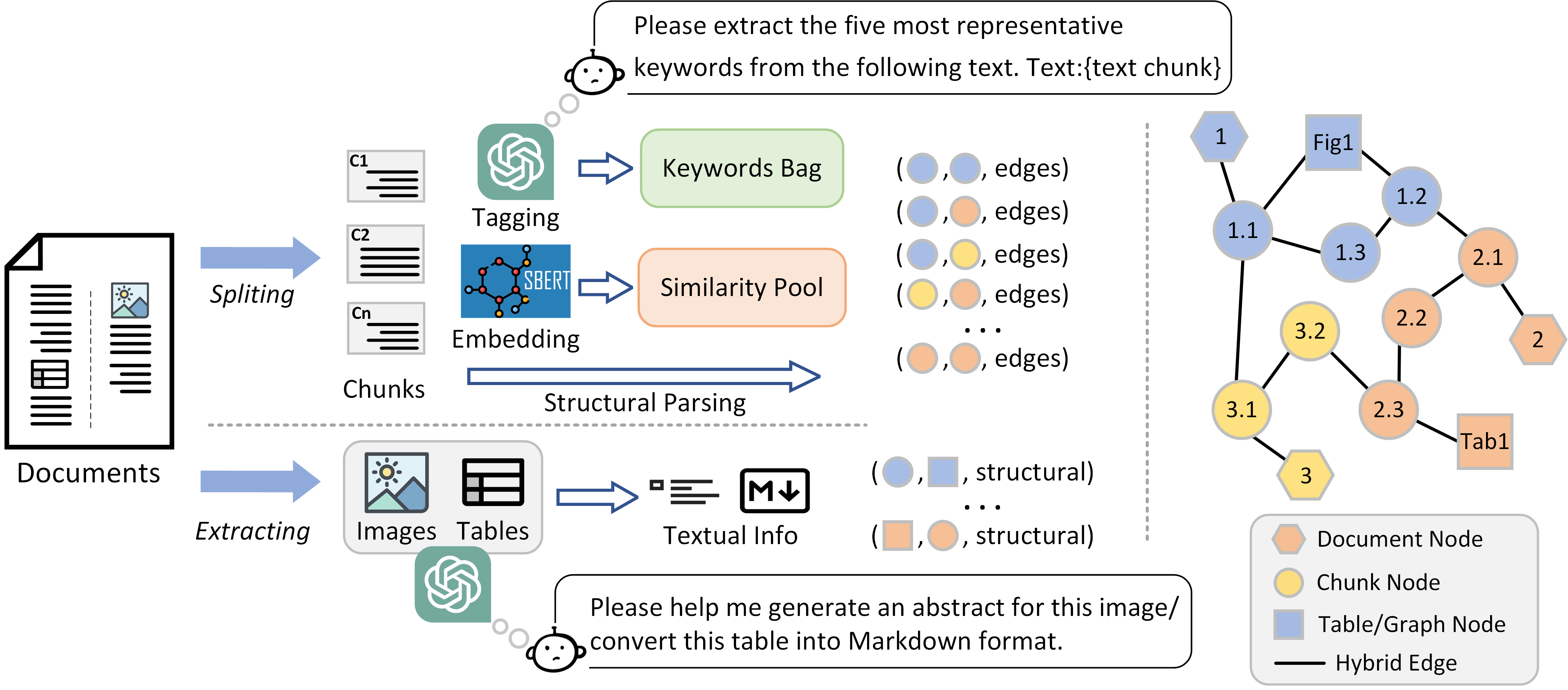}
    \caption{Illustration of the construction process of Chunk-Interaction Graph.}
    \label{fig:process}
\end{figure*}

\section{Method}

We propose a comprehensive document analysis method that constructs a Chunk-Interaction Graph by extracting interactions between chunks. By leveraging the graph structure and text attributes, this method facilitates the diffusion from high-similarity nodes to authentic supporting evidence, generating evidence chains with inherent reasoning logic. This approach significantly enhances the question-answering capabilities of LLMs.

\subsection{Chunk-Interaction Graph}

Isolating each paragraph during retrieval limits the retrieval and question-answering models from fully utilizing the contextual information for each paragraph. This presents the main challenge in  MDQA tasks. We believe that by fully considering the interactions between paragraphs, we can connect relevant paragraphs and enable them to influence each other during retrieval, thereby improving the coverage of supporting evidence.

A graph can integrate paragraphs into independent nodes and represent the inter-text relationships with topological connections between nodes. We design a Chunk-Interaction Graph (CIG) as an effective method for organizing numerous paragraphs across multiple documents. The construction process of the CIG is illustrated in Figure~\ref{fig:process}.

Fine-grained paragraphs can minimize irrelevant and noisy information in retrieval results. To achieve this, we first segment each document into chunks based on a fixed maximum size and punctuation. These chunks serve as nodes with text attributes in the graph, with their embeddings, topics, and original titles stored as additional meta-data. To accurately link and represent interactions between chunks using a straightforward graph topology, we design three types of edges with different weights to cover the relationships between chunk nodes: structural, semantic, and keyword. These edges can help the retrieval model overcome local limitations through multi-hop connections, building necessary bridges between the question and chunks that are not directly related.

\begin{figure*}[t]
  \centering
  \includegraphics[width=0.95\textwidth]{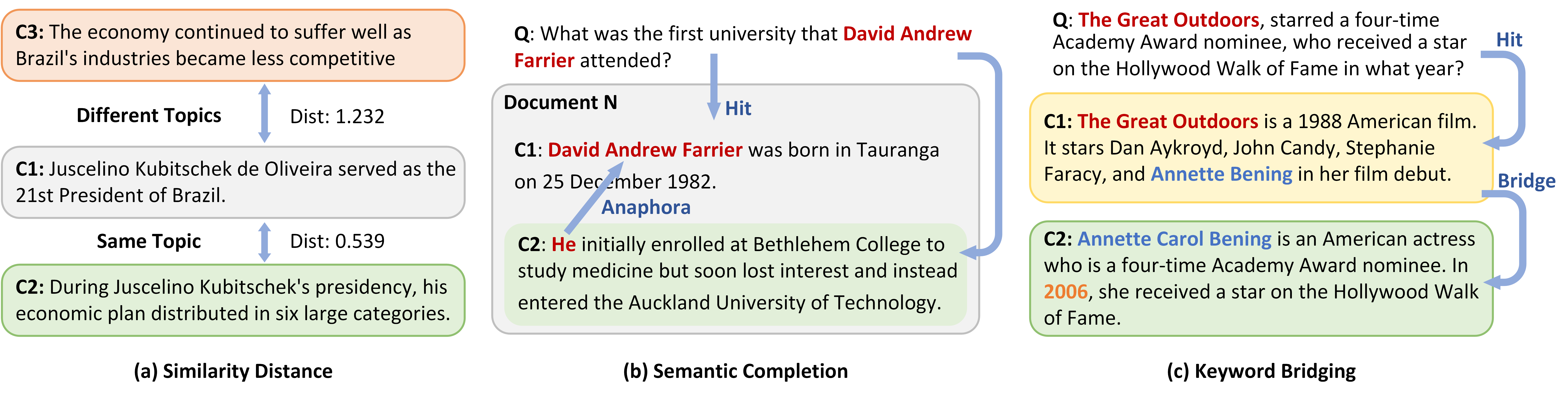}
  \caption{Illustration of three types of edges in Chunk-Interaction Graph. }
  \label{fig:edges}
\end{figure*}

One of the most commonly used methods for representing text correlations involves mapping chunks into a semantic space and measuring semantic similarity through distance calculations. Ideally, two chunks that describing the same topic, $C1$ and $C2$, should be close to each other in this semantic space, while an unrelated chunk $C3$ should be farther away, as shown in Figure~\ref{fig:edges}a. To establish the fundamental edge for linking chunks based on semantic similarity, we map all chunks into a shared embedding space to construct a similarity pool $P$. For each chunk, we filter out the top $k$ chunks with the highest semantic similarity to form semantic edges, with the similarity score as the weight. However, this method is limited by the completeness of the context within each chunk and the accuracy of the embedding model, making it insufficient to fully capture the interactions between chunks.

The structural relationships between chunks in the original document provide essential contextual information for accurately representing semantics. As illustrated in Figure~\ref{fig:edges}b, while $C1$ may not directly relate to the target question, it can correct the erroneous semantic representation of evidence chunk $C2$ by resolving ambiguity and anaphora issues in $C2$. Additionally, $C1$ can guide the reasoning process of LLM by supplying the missing logical information before and after the evidence chunk. To represent this structural interaction, we simply use the relative position of chunks within the document to establish  structural associations, thereby avoiding the complexities introduced by visual or spatial structures \cite{Gemelli2022Doc2GraphAT}. We start from the initial chunk of the document and add a structural edge to the adjacent chunk, sequentially connecting chunks within the same document as a chain. All structural edges are assigned a weight of 1.

Chunks from multiple documents, even if they discuss the same topic, often cover different aspects, making them hard to identify solely through semantic similarity. This challenge is prevalent in tasks that require the integration of multiple chunks to find an answer. As illustrated in Figure~\ref{fig:edges}c, the answer is contained in $C2$, which is not directly related to the target question $Q$, necessitating two-hop reasoning for retrieval. Keywords can succinctly represent the theme and main subject of a chunk. By identifying common keywords, we can cluster related chunks along a dimension different from semantics, constructing essential bridges between them. Furthermore, keyword association helps the retrieval model overcome local limitations, enabling cross-document knowledge retrieval and reasoning, which is a primary challenge in MDQA \cite{Pereira2022ViscondeMQ,Caciularu2023PeekAI}. To establish keyword edges correctly, we define keywords as the original phrases in a chunk that describe the main entities or themes. We extract keywords by prompting LLM with the template shown in \ref{app:extract keywords} and deploy the results as additional node attributes. All chunk keywords are compiled into a word bag $B$. By setting a keyword intersection threshold $T$, we filter out chunks discussing the same theme and construct keyword edges between the nodes. The edge weight corresponds to the size of the keyword intersection, with keywords also serving as attributes of the edge.  Notably, we do not include document titles as keywords, as titles describe the overall theme of the document, but cannot accurately summarize each chunk. Using title as a keyword might lead to an over-reliance on it during retrieval and lead to excessive intra-document linking, resulting in an overly dense graph.

In practical applications, documents typically contain various types of information, such as text, images, and structured data. Leveraging the advanced capabilities of multimodal LLMs, our approach can seamlessly extend to handle these varied documents. Specifically, we use multimodal LLMs to generate summaries of image contents\cite{ZHANG2024127530,Zhang2023LLaVAREV} and convert structured tables into Markdown format, which LLMs can effectively interpret \cite{Min2024ExploringTI,Zhang2023TableLlamaTO}. This additional information, along with the original multimodal data, is incorporated to build new nodes with specific attributes. These nodes are then connected to their adjacent text nodes based on their structural position within the document. This approach allows us to model various types of data into a unified CIG structure, enabling  consistent retrieval and reasoning methods across different data formats.

Based on the aforementioned design, CIG is defined as an ordered pair $CIG=\{\textit{V}, \textit{E}\}$, where $\textit{V}$ represents the set of nodes encompassing text chunks, images, and tables, and $\textit{E}$ represents the set of edges connecting these nodes. Specifically, each edge $e_{ij}$ contains three parameters: $W_{struc}$, $W_{sim}$, and $W_{keyword}$, which represent the three types of interactions and their respective weights.

\subsection{Evidence Chain Retriever}

The design of the CIG allows us to effectively mine and represent the interactions between numerous text pairs. To leverage the rich edge types, weights, and topological structure within the CIG and enhance retrieval accuracy, we design a retrieval model. This model iteratively approaches the supporting evidence, thereby completing the evidence chain.

To accurately retrieve all supporting evidence for a question $Q$ that requires knowledge from external documents, we employ a multi-path retrieval strategy to recall the evidence chains. The retriever begins by extracting a keyword set $Q_K$ with \ref{app:extract question keywords} to determine its topic and identify seed nodes $N_S$. The selection of the seed nodes involves a greedy search process that prioritizes the coverage of $Q_K$ and high semantic similarity. The retriever iteratively searches the CIG for the node that covers the most remaining keywords in $Q_K$ and has the highest semantic similarity to $Q$ as a seed node. All covered keywords will be removed until $Q_K$ is fully covered or no additional nodes match the keywords of $Q$. The aim of the greedy approach is to identify the smallest possible set of seed nodes, thereby minimizing the introduction of noise knowledge. We employ the same prompt template used during the construction of the CIG to extract the keywords and the same embedding model to calculate text similarity.

The retriever initiates its search from the seed nodes, iteratively identifying the supporting evidence of $Q$ and generating evidence chains to aid LLM in reasoning with the external knowledge. Starting from each seed node $N_{s,1}$, the retriever identifies the optimal neighboring node from all candidates and uses its text attributes to extend the path. The optimal neighboring node is defined as either the supporting evidence of $Q$ or a node that can guide the path to find a supporting evidence. To achieve this, we simultaneously capture previous paths, graph structure information, and features of neighboring nodes to enable more accurate retrieval. We construct a scoring model to evaluate each candidate node by integrating this information into a new representation.

First of all, we fine-tune a pretrained language model $f_u$ to construct a shared encoder, obtaining embeddings of the query, historical path, and potential text attributes, respectively:

\begin{equation}
E = f_u(T), \quad T \in \{Query, Path, Neighbour\}
\end{equation}

where $Query$ and $Neighbour$ are straightforward textual inputs, while $Path$ is the concatenation of the text attributes from the nodes in the traversed path. The output $E_i \in \mathbb{R}^D$ represents the embedding of each input, mapped to the same semantic space. This embedding is utilized to extract the latent semantic knowledge, enabling us to capture the impact of each candidate chunk on the evidence chain and the target question. 

Additionally, we use a 2-layer MLP $f_r$: $\mathbb{R}^3 \rightarrow \mathbb{R}^D$ to reconstruct the three types of interactions and their weights between chunks into a dense representation:

\begin{equation}
E_{Edge} = f_r(W_{struc}, W_{sim}, W_{keyword})
\end{equation}

where $W_{struc} \in \{0, 1\}$ indicates whether two chunks are adjacent in the original document, $W_{sim} \in \mathbb{R}^+$ represents the semantic similarity score, and $W_{keyword} \in \mathbb{N}$ denotes the number of common keywords between the chunks. Based on all the aforementioned representations, we employ a scoring function $f_n$ to evaluate the importance of neighboring nodes:

\begin{equation}
Score = f_n(E_{Query}, E_{Path}, E_{Neighbour}, E_{Edge})
\end{equation}

\begin{equation}
n_{i+1} = \arg\max_n (n|Score), n \in Neighbour(n_i)
\end{equation}

The function $f_n$: $\mathbb{R}^{4D} \rightarrow \mathbb{R}^1$ is also a 2-layer MLP. The score quantifies a candidate node's contribution to the completeness of the current path and its proximity to the evidence. During iteration $i$ of the retrieval process, we use $f_n$ to aggregate the scores of all neighboring nodes of $n_i$ and select the highest-scoring node as the next hop $n_{i+1}$. The selected node is iteratively added to the current path, updating the retrieval history, and its neighboring nodes become the new candidate nodes. This process continues until the maximum path length is reached.

To enhance the scoring model's capability to identify the evidence chunk, we guide the model's training by predicting which neighboring node will most quickly approach the evidence chunk. To achieve this goal, for each question, we start with an evidence node in the graph as the seed node and extract the shortest path between it and every other evidence node to form the training set. Nodes included in the shortest path are labeled as positive samples, while all other nodes are labeled as negative samples.

After completing all the retrieval processes, the chunks in each path are concatenated in retrival order as a chain to preserve the potential reasoning logic. These chains are then aggregated into a context as external knowledge to assist LLM in reasoning. The LLM is tasked with reasoning out the target question based solely on the provided retrieval content, without relying on any internal knowledge.

\section{Experiments}
\subsection{Dataset}

To evaluate the effectiveness of our method for MDQA tasks, we simulate the scenario of retrieving supporting evidence for the target question from a large corpus of documents. We extend the widely used QA datasets HotpotQA \cite{Yang2018HotpotQAAD}, 2WikiMQA \cite{Ho2020ConstructingAM}, IIRC \cite{Ferguson2020IIRCAD}, and MuSiQue \cite{Trivedi2021MM} all of which are built on Wikipedia. Specifically, we use the complete Wikipedia page relevant to each question as the potential corpus. Additionally, we add randomly selected pages as negative samples to supplement the Wikipedia topics of each question to a total of 12. All paragraphs are split into chunks, serving as the minimum text unit, thus creating a much larger and more complex retrieval scenario compared to the original dataset. Following the definition of SQUAD \cite{Rajpurkar2016SQuAD1Q}, we calculate the accuracy, EM, and F1 score of the answer given by LLM as evaluation metrics to assess performance.

\subsection{Baselines}

We compare the proposed IIER with previous state-of-the-art methods that utilize various retrieval paradigms: (1) TF-IDF, which uses word frequency as the feature to capture the relevance between the question and chunk, selecting the chunk with the most common keywords as evidence. (2) MDR \cite{Xiong2020AnsweringCO}, which iteratively encodes the question and historical path into a new query representation and implements maximum inner product search to select the relevant text chunk. (3) Adaptive-RAG \cite{Jeong2024AdaptiveRAGLT}, which first analyzes the complexity of the question and then dynamically adjusts the retrieval strategy to enhance the overall accuracy and efficiency of the QA system. (4) KGP \cite{Wang2023KnowledgeGP}, which fine-tunes LLMs to generate a hypothesis output based on the current retrieval context and calculates the similarity between the hypothesis output and potential chunks, employing a beam-search-like method to retrieve knowledge.

Additionally, we deploy the template shown in \ref{app: No Retrieval baseline} to prompt the LLM to answer the question without additional knowledge as the No Retrieval baseline. We use the template shown in \ref{app: QA} with the supporting evidence provided by the datasets to construct Golden baselines. For other baselines, we use the same prompt template \ref{app: QA} and backbone LLM to ensure fairness.

\subsection{Experiment Details}

We deploy all-mpnet-base-v2 as the foundational embedding model for constructing the CIG and for selecting seed nodes, utilizing cosine similarity to evaluate the semantic distance. In constructing the graph, the number of keywords extracted for each chunk is set to five. Each node forms similarity edges with the five nodes that have the highest similarity scores and keyword edge with nodes that share more than two keywords.

Based on the design of the evidence chain retriever, we fine-tune a model comprising roberta-base and two 2-layer MLPs to serve as the scoring model, guiding the retrieval of evidence on CIG. During the retrieval process, we set the maximum path length to five for all datasets. Therefore, for each question, we retrieve multiple evidence chains, each containing up to five chunks, starting from different seed nodes.

We use gpt-3.5-turbo as the backbone LLM to extract keywords and generate answers based on the retrieved external knowledge.

\begin{table*}[ht]
    \centering
    \begin{tabular}{l|ccc|ccc|ccc|ccc}
    \toprule
    \textbf{Method} & \multicolumn{3}{c}{\textbf{HotpotQA}} & \multicolumn{3}{c}{\textbf{2WikiMQA}} & \multicolumn{3}{c}{\textbf{IIRC}} & \multicolumn{3}{c}{\textbf{MuSiQue}} \\
    & Acc & F1 & EM & Acc & F1 & EM & Acc & F1 & EM & Acc & F1 & EM \\
    \hline
    No Retrieval & 57.4 & 39.9 & 30.1 & 45.5 & 19.9 & 32.5 & 35.6 & 16.5 & 17.3 & 35.0 & 15.9 & 5.5 \\
    \hline
    TF-IDF & 81.2 & 65.2 & 50.0 & 62.4 & 49.1 & 43.0 & 56.1 & 41.2 & 36.4 & 45.2 & 38.2 & 20.7 \\
    \hline
    MDR & 80.4 & 66.0 & 52.0 & 66.0 & 49.8 & 42.1 & 61.3 & 45.3 & 38.7 & 46.3 & 42.9 & 21.3 \\
    \hline
    Adaptive-RAG & 87.5 & 71.9 & 56.9 & 66.7 & 52.5 & 46.7 & 60.3 & \textbf{45.8} & \textbf{40.4} & 50.4 & 45.0 & 25.2 \\
    KGP & 86.7 & 70.0 & 54.5 & 69.8 & \textbf{55.5} & 45.0 & 59.8 & 44.3 & 38.3 & 60.5 & 48.6 & 34.2 \\
    IIER (ours) & \textbf{93.5} & \textbf{76.4} & \textbf{63.6} & \textbf{84.8} & 53.0 & \textbf{47.3} & \textbf{64.8} & 45.1 & 38.6 & \textbf{62.5} & \textbf{49.5} & \textbf{35.0} \\ 
    \hline
    Golden & 97.8 & 76.7 & 64.5 & 92.3 & 59.4 & 52.5 & 85.7 & 56.0 & 42.9 & 76.7 & 54.6 & 31.9 \\
    \bottomrule
    \end{tabular}
\caption{Performance of IIER and baselines on MDQA datasets. The best results are in bold.}
\label{Tab Main Result}
\end{table*}

\subsection{Main Results}

The main results of IIER and the baselines on MDQA datasets are presented in Table~\ref{Tab Main Result}. IIER achieves the best accuracy across all datasets by effectively extracting and analyzing chunk interactions to construct evidence chains. Specifically, IIER shows significant accuracy improvements of 15\% on 2WikiMQA and 6.0\% on HotpotQA compared to the baselines. For IIRC and MuSiQue, which require more stringent multi-hop reasoning and retrieval accuracy, IIER also improves the accuracy by 3.5\% and 2.0\%, respectively.

Among the baselines, MDR and Adaptive-RAG both enhance retrieval by utilizing the previous path and process information for next hop selection. However, they treat each potential chunk as an isolated entity and calculate the selection probability or word frequency score independently for each chunk. KGP transforms documents into graph structures based on keywords, but the TAGME \cite{Ferragina2010TAGMEOA} they employ can only recognize entities contained in Wikipedia. This approach overlooks the semantic information of the chunks and adds excessively noisy links to the graph, making it challenging to accurately capture associations between chunks. In contrast, IIER addresses this issue by designing multiple types of interactions to comprehensively cover the potential semantic and logical connections between chunks. By constructing CIG based on these connections, IIER enables the simultaneous creation of multiple reasoning chains to retrieve supporting evidence. These mechanisms provide IIER with superior cross-document knowledge reasoning capabilities, enabling it to achieve outstanding performance in MDQA tasks. Furthermore, compared to Adaptive-RAG and KGP, which utilize LLM to guide retrieval iteratively, IIER treats LLM as a plug-and-play module to read external knowledge and answer questions. This design offers IIER higher retrieval efficiency and lower transfer overhead.

In the 2WikiMQA and IIRC datasets, IIER achieves the highest accuracy but does not obtain the best F1 score or EM. This discrepancy arises because the correct answers provided in datasets are often short spans containing only the target entity or a simple yes/no response. Despite instructions to limit the output, the LLM tends to generate complete sentences, such as ``Yes, both in Iran.", which results in a lower score. Therefore, while the F1 score and EM follow the same trend as accuracy, they do not exhibit a strict linear relationship. We use accuracy as the evaluation metric in all subsequent experiments to assess the model's performance.

\subsection{Effectiveness of Evidence Chain}

To evaluate the effectiveness of constructing evidence chains in retrieving supporting evidence and assisting LLM in reasoning, we design two experiments to analyze the impact of evidence chain construction and the scope of the evidence chain on IIER performance.

\begin{table}[h]
    \centering
    \begin{tabularx}{\columnwidth}{lcccc}
    \toprule
    \textbf{Method} & \textbf{HotpotQA} & \textbf{2WMQA} & \textbf{IIRC} & \textbf{MuSiQue} \\
    \hline
    IIER & & & & \\
    w/ Shuffle & 89.6 & 79.0 & 57.5 & 56.1 \\ 
    w/ Iterative & 92.1 & 83.9 & 58.5 & 57.4 \\
    w/ Chain & \textbf{93.5} & \textbf{84.8} & \textbf{64.8} & \textbf{62.5} \\
    \bottomrule
    \end{tabularx}
\caption{Performace of using different formats to concatenate retrieval results on MDQA datasets.}
\label{Tab Chain Construction}
\end{table}

\subsubsection{Impact of Evidence Chain Construction}

To demonstrate the effectiveness of the chain structure, we use three formats to organize the retrieved chunks and add them to the prompts for LLM. First, we randomly shuffle all retrieved chunks to simulate the simple strategy in Naive RAG \cite{Gao2023RetrievalAugmentedGF}. Second, we also concatenate the chunks retrieved in the same iteration and arrange them in iteration order to simulate the iterative strategy. The results on four datasets are shown in Table~\ref{Tab Chain Construction}. The results indicate that the iterative format slightly outperforms the random format, suggesting that the order of chunks is crucial for maintaining the coherence of logic and context, aiding LLM in reasoning. Furthermore, the chain format significantly enhances IIER's performance. This not only demonstrates that our designed CIG can effectively model the interactions between chunks, but also shows that our retriever can correctly guide the search direction. Consequently,  IIER can provide complete logic and context while identifying the supporting evidence.

\begin{figure*}[t]
  \centering
  \begin{minipage}[t]{0.45\linewidth}
      \centering
      \includegraphics[width=\linewidth]{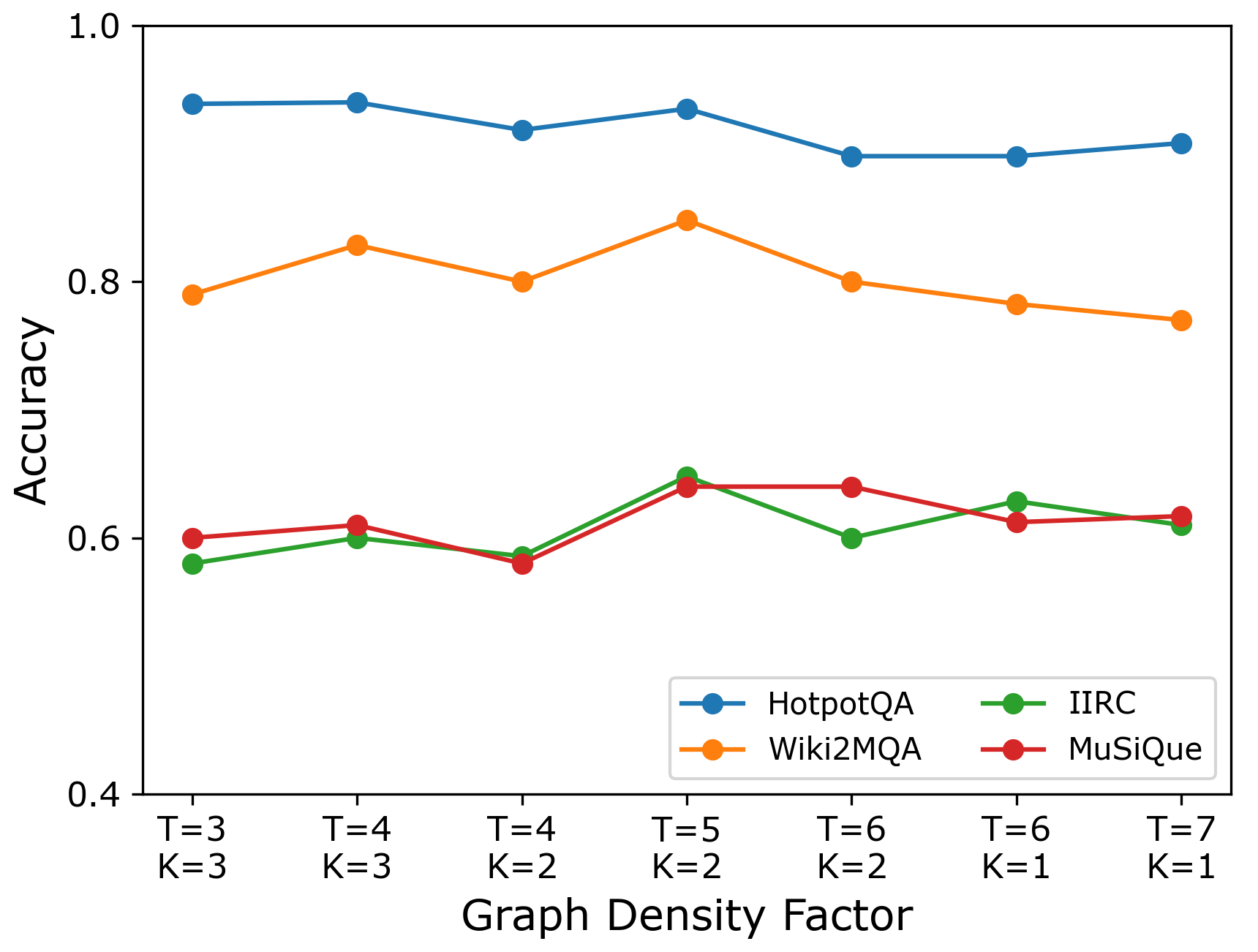}
  \end{minipage}
  \begin{minipage}[t]{0.45\linewidth}
      \centering
      \includegraphics[width=\linewidth]{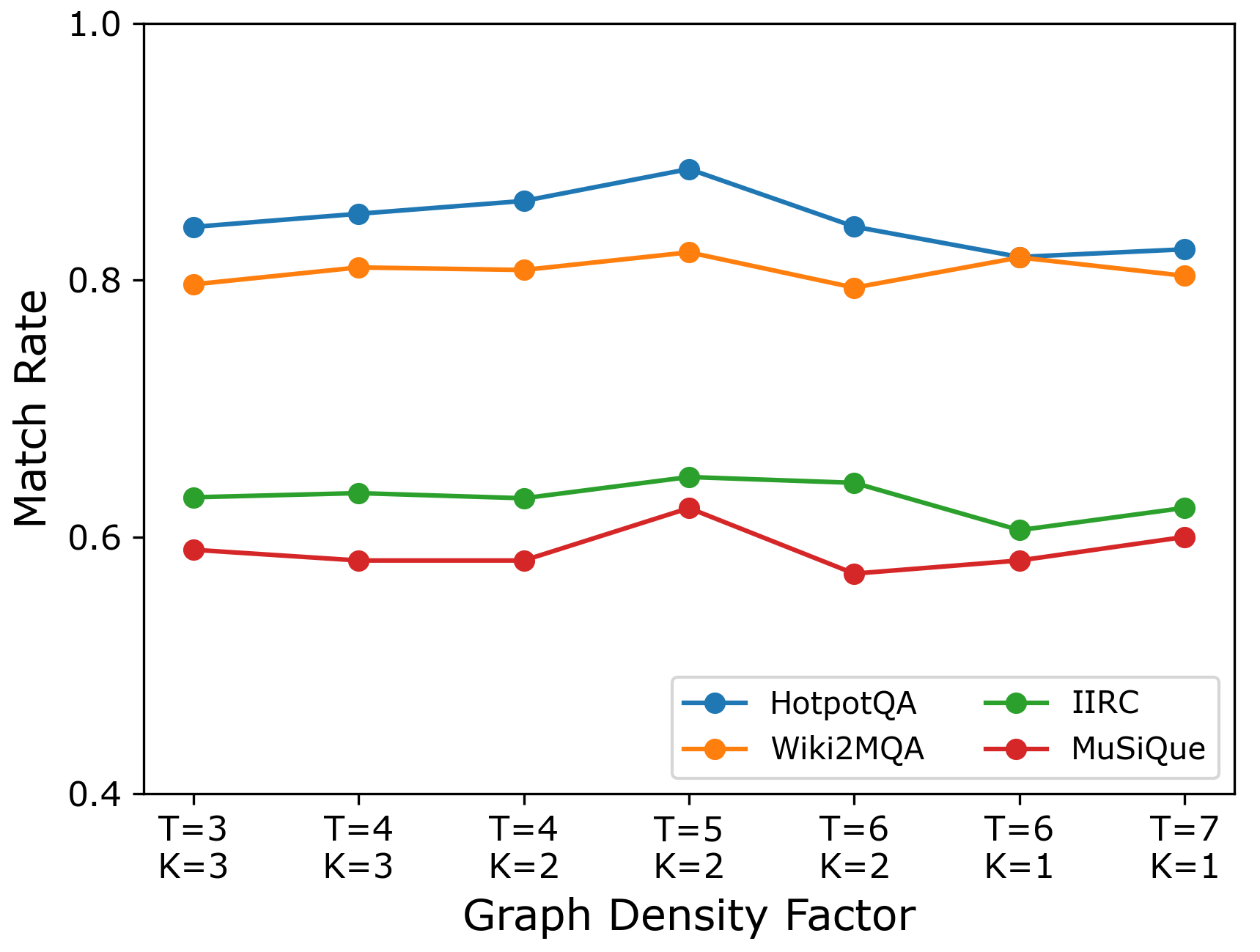}
  \end{minipage}
  \caption{Performance of IIER with different graph density on MDQA datasets.}
  \label{fig:density}
\end{figure*}

\subsubsection{Impact of Evidence Chain Scope}
We control the search scope of the retriever on CIG by setting different maximum lengths for the evidence chain. Specifically, when the maximum length is set to 1, the retriever can only use the seed nodes as retrieval results. The results, as shown in Table~\ref{Tab Chain Scope}, indicate that using only the seed nodes as external knowledge yields the worst performance. This demonstrates that independently retrieving chunks is insufficient to provide all the information needed to answer the question due to the lack of context. When the chain is short, extending its length helps the retriever utilize the interactions in CIG to expand its search scope and find more supporting evidence globally, thereby improving the accuracy of LLM reasoning. However, once the chain's length reaches a threshold, the retriever has already identified all supporting evidence, and further extending the length only introduces limited additional contextual information, causing the accuracy to plateau. This is consistent with the results shown in Table~\ref{Tab Chain Scope}, where the performance of IIER improves and then remains stable as the chain length increases.

\begin{table}[t]
  \centering
  \begin{tabularx}{\columnwidth}{lcccc}
  \toprule
  \textbf{Method} & \textbf{HotpotQA} & \textbf{2WikiMQA} & \textbf{IIRC} & \textbf{MuSiQue} \\
  \hline
  IIER & & & & \\
  w/ 1-step & 89.0 & 82.1 & 55.6 & 58.6 \\
  w/ 3-step & 90.9 & 83.0 & 61.6 & 59.0 \\
  w/ 5-step & 93.5 & \textbf{84.8} & 64.8 & 62.5 \\ 
  w/ 7-step & \textbf{94.0} & 84.2 & \textbf{65.9} & \textbf{63.1} \\
  \bottomrule
  \end{tabularx}
\caption{Performance of IIER with different evidence chain scope on MDQA datasets.}
\label{Tab Chain Scope}
\end{table}

\subsection{Impact of Chunk-Interaction Graph}

We study the impact of the graph density on retrieval and reasoning performance by adjusting the hyperparameters of CIG to construct graphs with varying densities. The number of structural edges in CIG is determined by the documents. Therefore, we control the graph's density by modifying the number of most relevant chunks $T$ for constructing semantic edges and the minimum keyword threshold $K$ for constructing keyword edges. 

We conduct experiments on all four datasets and evaluate the accuracy of IIER and the match rate of supporting evidence. The results are shown in Figure~\ref{fig:density}.

The results indicate that when the graph density is low, the match rate of IIER remains stable across all datasets, while the accuracy shows a fluctuating upward trend. This occurs because fine-tuning the retriever enables it to accurately search for supporting evidence even when limited correct paths are available. These relevant pieces of information can support answering the question, but the lack of context information restricts the reasoning capabilities of the LLM. As the number of edges between nodes increases, more paths connecting seed nodes and supporting evidence appear in the graph. The retriever can discover paths with stronger semantic relevance and shorter lengths to replace the suboptimal paths, thereby enabling LLM to further enhance its understanding and reasoning. However, when the density surpasses a certain threshold, the excessive potential paths introduce noise, confusing the model and making it difficult to select the optimal neighbor in each iteration. Consequently, both the accuracy and match rate slightly decrease. Additionally, higher density necessitates evaluating more nodes in each iteration, increasing the computational overhead and reducing retrieval efficiency. Therefore, in practical applications, it is necessary to adjust and select a moderate density based on the retrieval task requirements to balance efficiency and accuracy.

\section{Conclusion}
Completing the MDQA tasks requires retrieving all supporting evidence from documents across different sources to achieve cross-passage complex reasoning, posing a significant challenge for the RAG paradigm. We recognize that isolated paragraph processing leads to context loss, ambiguous references, and other issues that hinder retrieval and reasoning. Therefore, we propose IIER, which enhances retrieval and aids LLM reasoning through inter-chunk interactions.    IIER extracts structural, semantic, and keyword associations between chunks, unifying multi-source documents into a Chunk-Interaction Graph to bridge related chunks and context. To leverage this information for retrieving supporting evidence, we design a graph-based evidence chain retriever. This retriever iteratively selects the optimal path by leveraging topological and semantic information to approach the supporting evidence. It constructs the retrieval results into evidence chains with complete context and reasoning logic to assist the LLM in answering the target question. Experiments demonstrate the superiority of IIER compared to the baselines, highlighting the effectiveness of CIG and the evidence chain in enhancing retrieval and reasoning.

\appendix
\section{Prompt Templates Throughout the Work}

We display the prompt templates used in IIER process, where all \{question text\} indicates the target question, \{chunk text\} indicates the paragraphs in the documents, and \{context text\} indicates the retrieved knowledge.

\subsection{Prompt Template for Extracting Keywords from Questions} 
\label{app:extract question keywords}

To cover all relevant topics of the question for seed node selection, we do not limit the number of keywords extracted from the question. The prompt template for extracting keywords from the questions is shown as follows:

\begin{minipage}{0.92\columnwidth}
    \centering
    \vspace{1mm}
    \begin{tcolorbox}[title= Prompt template for extracting keywords from questions]
        \small
        \textbf{Instruction:} Please select all the topics and keywords covered in the following query and return them as a list with the keywords separated by commas.  \\

        \textbf{Example:}\\
        \textbf{Question A:} When did the people who captured Malakoff come to the region where Philipsburg is located?\\
        \textbf{Answer A:}\\
        \textnormal{\textup{['Philipsburg', 'Malakoff']}}
        \\

        \textbf{Question B:} When was the first establishment that McDonaldization is named after, open in the country Horndean is located?\\
        \textbf{Answer B:}\\
        \textnormal{\textup{['McDonaldization', 'Horndean']}}
        \\

        \textbf{Question:} \\
        \{question text\} \\
        \textbf{Answer:} 
    \end{tcolorbox}
    \vspace{1mm}
\end{minipage}

\subsection{Prompt Template for Extracting Keywords from Chunks} 
\label{app:extract keywords}

The prompt template for extracting keywords from the paragraphs of the documents is shown as follows:

\begin{minipage}{0.92\columnwidth}
    \centering
    \vspace{1mm}
    \begin{tcolorbox}[title= Prompt template for extracting keywords from chunks]
        \small
        \textbf{Instruction:} Please extract the five most representative keywords from the following text and return them as a list with the keywords separated by commas. \\

        \textbf{Example:} \\
        \textbf{Text:} John Cecil, 6th Earl of Exeter (15 May 1674 – 24 December 1721), known as Lord Burleigh from 1678 to 1700, was a British peer and Member of Parliament. He was the son of John Cecil, 5th Earl of Exeter, and Anne Cavendish.\\
        \textbf{Answer:}\\
        \textnormal{\textup{['John Cecil, 6th Earl of Exeter', 'Lord Burleigh', 'British peer', 'Member of Parliament', 'John Cecil, 5th Earl of Exeter']}}
        \\

        \textbf{Text:} \\
        \{chunk text\} \\
        \textbf{Answer:} 
    \end{tcolorbox}
    \vspace{1mm}
\end{minipage}

\subsection{Prompt Template for Question Answering}
\label{app: QA}

To avoid introducing noisy knowledge, we do not provide fewshot information and only use retrieved knowledge as input. The prompt template for question answering is shown as follows:

\begin{minipage}{0.92\columnwidth}
    \centering
    \vspace{1mm}
    \begin{tcolorbox}[title= Prompt template for question answering]
        \small
        \textbf{Instruction:} Given the following question and contexts, generate a final answer to the question. Please answer in less than 6 words. \\

        \textbf{Question:} \\
        \{question text\} \\
        \textbf{Context:} \\
        \{context text\} \\
        \dots \\
        \textbf{Answer:}
    \end{tcolorbox}
    \vspace{1mm}
\end{minipage}

\subsection{Prompt Template for No Retrieval Baseline}
\label{app: No Retrieval baseline}

We directly use the original questions provided by the dataset to prompt LLM to generate answers as the No Retrieval baseline. To avoid interference, we do not provide few-shot information. The prompt template for No Retrieval baseline is shown as follows:

\begin{minipage}{0.92\columnwidth}
    \centering
    \vspace{1mm}
    \begin{tcolorbox}[title= Prompt template for No Retrieval baseline]
        \small
        \textbf{Instruction:} Given the following question, generate an answer to the question. Please answer in less than 6 words. \\

        \textbf{Question:} \\
        \{question text\} \\
        \textbf{Answer:} 
    \end{tcolorbox}
    \vspace{1mm}
\end{minipage}

\bibliographystyle{elsarticle-num} 
\bibliography{ref}

\end{document}